\documentclass[10pt,twocolumn,letterpaper]{article}

\usepackage{iccv}
\usepackage{times}
\usepackage{epsfig}
\usepackage{graphicx}
\usepackage{amsmath}
\usepackage{amssymb}
\usepackage{algorithmic}
\usepackage[hang]{subfigure}
\usepackage{textcomp}
\usepackage[ruled]{algorithm2e}
\usepackage{multirow}
\usepackage{diagbox}
\usepackage{verbatim}
\usepackage{xcolor}
\usepackage{adjustbox}
\usepackage{caption}
\usepackage[symbol]{footmisc}

\usepackage[pagebackref=true,breaklinks=true,letterpaper=true,colorlinks,bookmarks=false]{hyperref}

\iccvfinalcopy 

\def\iccvPaperID{6028} 

\ificcvfinal\fi

\begin{document}

\title{Restore from Restored: Single-image Inpainting}

\author{Eunhye Lee\footnotemark[1], Jeongmu Kim\footnotemark[1], Jisu Kim\footnotemark[1] and Tae Hyun Kim\footnotemark[2]\\
Dept. of Computer Science, Hanyang University\\
Seoul, South Korea\\
{\tt\small $\{$dldms1345, jmkim1503, atat1270, taehyunkim$\}$@hanyang.ac.kr}
}

\maketitle
\ificcvfinal\thispagestyle{empty}\fi

\footnotetext[1]{Eunhye Lee, Jeongmu Kim and Jisu Kim are co-first authors.}
\footnotetext[2]{Tae Hyun Kim is corresponding author.}
\begin{abstract}
Recent image inpainting methods show promising results due to the power of deep learning, which can explore external information available from a large training dataset.
However, many state-of-the-art inpainting networks are still limited in exploiting internal information available in the given input image at test time.
To mitigate this problem, we present a novel and efficient self-supervised fine-tuning algorithm that can adapt the parameters of fully pre-trained inpainting networks without using ground-truth target images.
We update the parameters of the pre-trained state-of-the-art inpainting networks by utilizing existing self-similar patches within the given input image without changing network architecture and improve the inpainting quality by a large margin.
Qualitative and quantitative experimental results demonstrate the superiority of the proposed algorithm, and we achieve state-of-the-art inpainting results on publicly available numerous benchmark datasets.
\end{abstract}

\section{Introduction}

\begin{figure}[t]
    \centering
    \begin{adjustbox}{max width=1.0\linewidth,center}
    \includegraphics{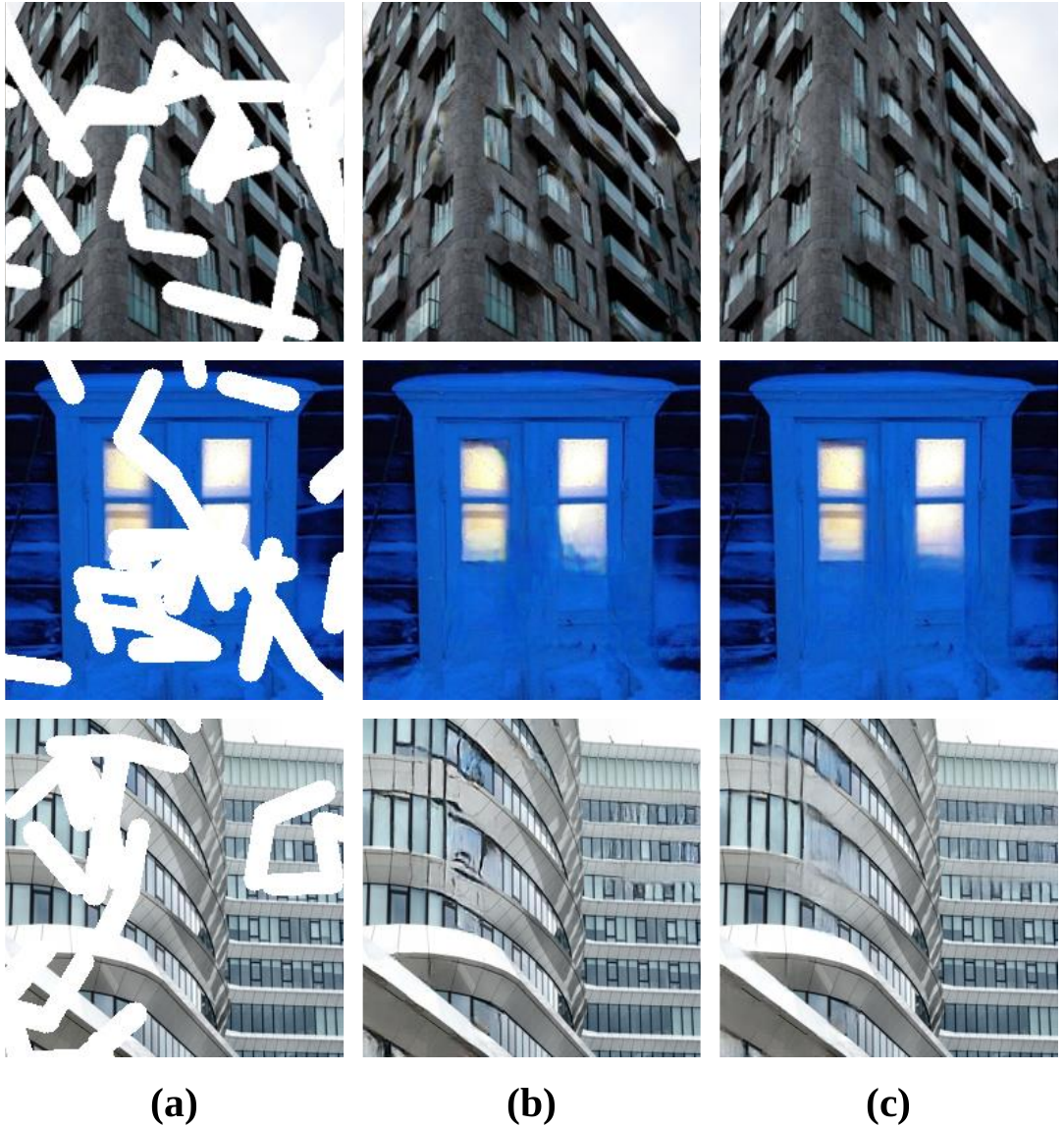}
    \end{adjustbox}
    \caption{\small
    Results of our approach.
    (a) Masked input images.
    (b) Initially restored results from pre-trained inpainting networks in the order of GatedConv\cite{yu2019free}, EdgeConnect\cite{nazeri2019edgeconnect}, and GMCNN\cite{wang2018image}.
    (c) Results of fine-tuned versions of GatedConv\cite{yu2019free}, EdgeConnect\cite{nazeri2019edgeconnect}, and GMCNN\cite{wang2018image} using the proposed learning algorithm. Restored regions are more natural and consistent with other regions than the initial results.}
    \label{fig:intro}
\end{figure}

\label{sec:introduction}
Image inpainting, a post image processing technique that removes unnecessary parts of images, such as subtitles and obstacles, and fills these areas with new colors and textures, has extensive applications in areas of photo editing, restoration, and even video editing~\cite{barnes2009patchmatch, levin2004seamless, yu2019free, newson2014video, kim2019deep}.
The main goal of inpainting is to generate plausible images that are semantically consistent with the overall context and to fill the missing area to be continuous with the surrounding area.

Many traditional inpainting approaches based on the diffusion technique compute pixel values of the missing area using pixel values in the surrounding area~\cite{bertalmio2000image, ballester2001filling, levin2003learning}.
Although diffusion-based approaches are effective in recovering small areas (e.g., subtitle removal), they produce blurry results when the missing area is large (e.g., object removal).
As an alternative, patch-match-based methods have been proposed to locate similar patches within the given image and copy intensity values of these similar patches into the missing area~\cite{liang2001real, hays2007scene, barnes2009patchmatch}.
This method can fill the missing area with realistic image contents using information available within the input image.
However, it is difficult to restore faces or complex landscapes that require information outside the given image, and it creates irrelevant structures due to the lack of image context inference capabilities.

The development of deep learning technologies has remarkably improved the performance of image inpainting techniques and made it possible for neural approaches to reconstruct considerably challenging images.
First, context encoder pioneered the use of deep neural networks in the image inpainting task and introduced an architecture composed of convolutional encoder and decoder with adversarial loss~\cite{pathak2016context, goodfellow2014generative}.
The encoder computes a latent feature representation from the input image and the decoder uses the feature representation to restore pixels in the missing area, consequently inferring and reconstructing the overall structure of the input image.
The output demonstrates a realistic texture as it approaches the distribution of the real image through the adversarial loss.
Follow-up studies carry out various attempts while maintaining encoder-decoder structures with discriminators, presented in the work of context encoders~\cite{pathak2016context}.
Among them, attention modules are exploited to generate realistic textures~\cite{yu2018generative, ren2019structureflow, zeng2019learning, li2020recurrent}.
These attention modules allow searching non-local regions to utilize information among similar structures in the given input image and fill pixel values of missing areas.
Some studies perform structural restoration before carrying out the inpainting to clarify the boundary of the missing hole~\cite{ren2019structureflow, nazeri2019edgeconnect, xiong2019foreground}.

Although these neural approaches trained in a supervised manner with a large external training dataset for the inpainting task have shown promising and satisfying results, they have difficulties in fully utilizing the information within given test images at the test phase.
Thus, we propose a new learning approach to utilize internal statistics available within the given input test image such as patch-recurrence to overcome this limitation.
Patch-recurrence is a property that many similar patches are existing within a single natural image and numerous image restoration techniques utilize these recurring patches to improve the restoration quality~\cite{michaeli2014blind, glasner2009super, huang2015single}.
In particular, there are several attempts to train the restoration networks for super-resolution and denoising tasks using only internal statistics without the ground-truth clean images~\cite{shocher2018zero, krull2019noise2void, batson2019noise2self}, and these self-supervised approaches have shown satisfactory results.
However, these self-supervision-based approaches fail to exert the power of deep learning through large external datasets.

We demonstrate that inpainting networks can be also trained in a self-supervised manner for the first time and develop a new learning algorithm that combines supervised and self-supervised approaches to benefit from both the external dataset and the given input test image in this work.
This approach improves the performance of existing state-of-the-art inpainting networks by simply updating network parameters using the internal information (i.e., repeating structure/texture and color distribution) available from the given input image at test time.
Specifically, we use parameters of fully pre-trained inpainting networks on the large external dataset as initial values.
We then fine-tune the network parameters in a self-supervised manner at the test stage by exploiting self-similar patches within the input test image and produce improved results as shown in Figure~\ref{fig:intro}.
Note that broken lines and crushed edges of the initial result images are clearly restored and the original shapes are properly predicted with the proposed fine-tuning algorithm.
Our learning method is not restricted to specific network architecture and can be applied to various conventional networks.
Moreover, our approach can easily upgrade the parameters of the conventional inpainting networks without changing their original architecture.
The main contributions of this study are summarized as follows:
\begin{itemize}
	\item A novel self-supervised fine-tuning algorithm that automatically takes advantage of recurring patches within the test image is proposed.
	\item Superior inpainting results on benchmark datasets are achieved by utilizing both internal and external datasets.
	\item Our approach can be applied to various inpainting networks without modifying their original network architecture and loss functions.
\end{itemize}

\section{Related Work}
\label{sec:related}

\subsection{Image inpainting}
Recently, deep learning methods have successfully been used in many computer vision tasks, and several learning-based approaches have been proposed for inpainting.
Context encoder~\cite{pathak2016context} introduces an encoder–decoder network based on convolutional neural network (CNN) to extract semantics of the image and employs GAN~\cite{goodfellow2014generative} architecture to generate realistic details.
It overcomes the limitation of traditional methods that struggle to create unique content, such as human faces or complex scenes because they fill the missing area using only the information within the given input~\cite{bertalmio2000image,ballester2001filling,telea2004image,Darabi12:ImageMelding12,barnes2009patchmatch,huang2014image}.
Iizuka et al.~\cite{iizuka2017globally} removed the pooling layer and employed dilated convolutions~\cite{yu2015dilated} in the neural networks to solve the problem of the context encoder that generates blurry outputs because of the reduction of the resolution during the encoding stage.
Also, they used two discriminators to maintain continuity in and out of the hole.
The local one focus on the missing area and the global one assess the entire image.
Yu et al.~\cite{yu2018generative} analyzed that CNN creates visual artifacts because of the inability to exploit image contents far from the hole.
Thus, contextual attention module is proposed to overcome this limitation.
Contextual attention module scores the similarity of background patches with pixels of the missing area which are estimated in advance by a coarse network.
Then, patches with high similarity scores are used to reconstruct the missing area.
Generative multi-column CNN~\cite{wang2018image} uses parallel networks with different receptive field sizes to prevent the error propagation caused by the coarse and fine networks that are connected in series.
It also uses the reconstruction loss that gives different weights to the pixels in the hole considering the distance from the boundary to solve the spatial-variant constraints.
Liu et al.~\cite{liu2018image} proposed the use of partial convolution to learn various forms of masks.
Previous methods could not be utilized in practical cases where images were damaged in various forms because they were typically learned with squared masks.
Partial convolution updates the mask during the inference and renormalizes weights of the masked area every layer to propagate only valid information of the input.
In a similar approach by Yu et al.~\cite{yu2019free}, gated convolution automatically finds a proper mask for the input and considers the gating value calculated across a spatial location of every layer and channel.
EdgeConnect~\cite{nazeri2019edgeconnect} introduces a two-stage inpainting network model that first generates the edge map with canny edge detector~\cite{canny1986computational} and then recovers the whole image.
However, this model is limited by its inability to utilize additional useful information, such as image color.
StructureFlow~\cite{ren2019structureflow} uses a smooth image while preserving the edges obtained via RTV~\cite{xu2012structure} instead of the edge map and employs appearance flow module~\cite{zhou2016view} to achieve an attention effect for realistic texture.
Li et al.~\cite{li2020recurrent} presented recurrent feature reasoning (RFR) network and knowledge consistent attention (KCA) module to cover large holes.
They reduce the region of the hole progressively using partial convolution and subsequently merge the inpainting results for each hole.
At each iteration, KCA module recurrently calculates the attention score of the missing hole based on the attention score computed in the previous step.

\subsection{Self-supervised learning for restoration}
The self-supervised approach for image restoration primarily refers to the training of the neural networks for the restoration of corrupted images using only information from the input images themselves without corresponding clean images~\cite{lehtinen2018noise2noise, krull2019noise2void, batson2019noise2self, shocher2018zero}.
For the super-resolution task, zero-shot SR (ZSSR)~\cite{shocher2018zero} trains a small and image-specific CNN during test time and utilizes internal statistics such as patch-recurrence within a given input image.
However, these methods can only utilize internal data because they train the network from random scratch at the test time.

We present that the self-supervised approach can be applied to image inpainting to fine-tune parameters of networks during test time by exploiting internal statics of a given masked input image.
Moreover, our method exploits the advantage of both external and internal data by fine-tuning the fully pre-trained networks like~\cite{park2020fast,lee2020restore} to overcome the limitation of the self-supervised approach that fails to exert the power of deep learning via large external datasets.

\section{Proposed Method}
In this section, we introduce a simple, yet effective self-supervised learning approach to adapt the parameters of pre-trained inpainting networks to the given specific input test image.

\subsection{Patch-recurrence for inpainting}
The internal information available within the input image is very important for the image inpainting task and should be considered to generate semantically consistent and realistic results when filling missing areas.

Images generally demonstrate the property of patch recurrence, which enables the repeated presentation of many identical or similar patches in a single image.
The information of similar patches inside the image is used in past patch-match-based methods~\cite{Darabi12:ImageMelding12, hays2007scene, barnes2009patchmatch, huang2014image} and recent deep-learning methods with the attention mechanism~\cite{yu2018generative,ren2019structureflow,li2020recurrent}.
Patch-match-based approaches find similar patches within the target image to fill the hole and paste them.
By comparison, the attention module~\cite{yu2018generative, li2020recurrent} and appearance flow~\cite{ren2019structureflow} in deep-learning approaches find similar patches/features in the given input image corresponding to the target (missing) area, and resulting attention maps are used to complete the texture within that area.
These methods improve the inpainting results by exploiting the internal information during test time but require an additional network module (e.g., non-local operator) to compute attention maps and are inefficient because they require high computational costs to measure similarity among features at every pixel location.
Specifically, $M \times N$ comparisons are required when the number of pixel values within missing and non-missing areas is $N$ and $M$, respectively, but only a few matches of them produce meaningful high-similarity scores.

Therefore, we introduce a new approach that can fully utilize recurring patches in the given input image during the test phase without explicit similarity comparisons (i.e., patch-match) and also present a new fine-tuning method that enhances the quality of inpainting results of existing networks in a self-supervised manner.

\subsection{Patch-match-based inpainting without patch-match}

\begin{figure*}
    \centering
    \includegraphics[width=1.0\linewidth]{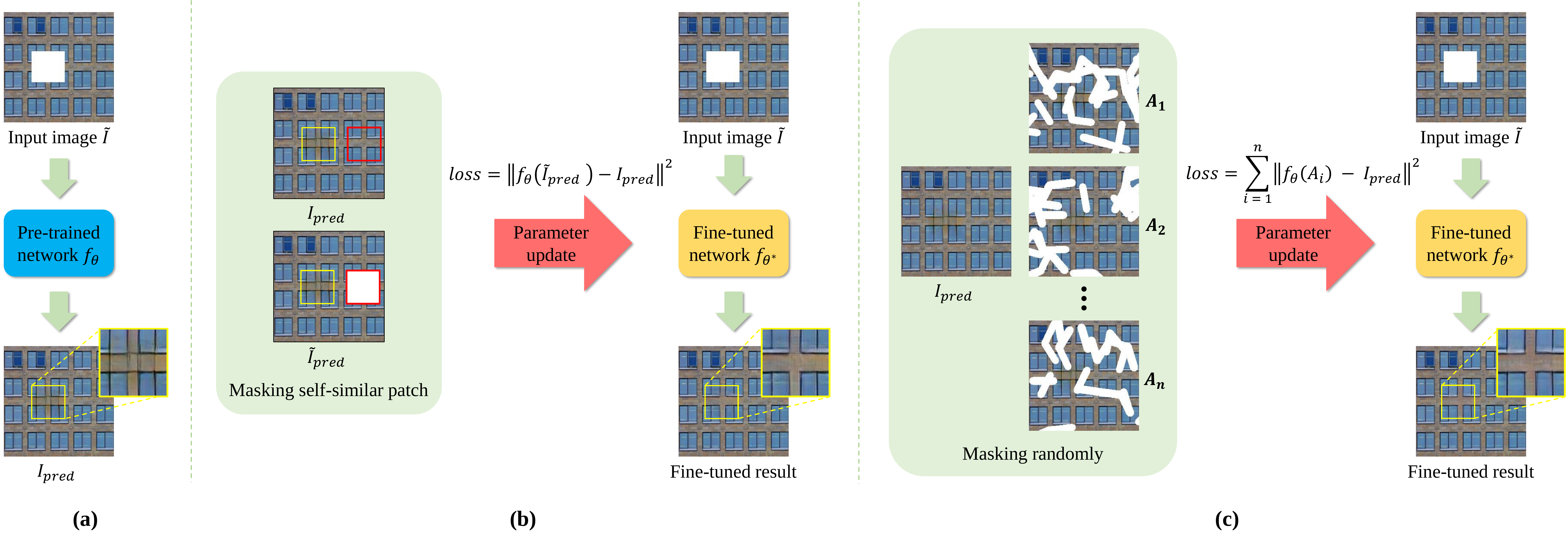}
    \caption{\small Inpainting results by EdgeConnect~\cite{nazeri2019edgeconnect} before and after fine-tuning.
    (a) Fully pre-trained EdgeConnect~\cite{nazeri2019edgeconnect} produces artifacts.
    (b) Fine-tuned EdgeConnect~\cite{nazeri2019edgeconnect} using a corresponding patch (red box) within the input can remove artifacts.
    (c) Fine-tuned EdgeConnect~\cite{nazeri2019edgeconnect} using randomly masked input can also remove artifacts without finding correspondences.
    }
    \label{fig:patchrecurrence}
\end{figure*}

Networks in existing learning-based image inpainting methods learn how to reconstruct the missing area using a large number of training images during the training phase.
However, the input image given at the test phase may include unique and unseen information, such as unusual structures, textures, and colors, which are rare in the training dataset although a large external dataset is available for training.
In this case, the pre-trained network may generate an unnatural output incompatible with the specific pattern in the background.
Therefore, we propose a new inpainting approach to exploit the specific information available within the input test image by utilizing multiple similar patches in the input and further improve the pre-trained network by adapting network parameters to the specific input image.

Figure \ref{fig:patchrecurrence} illustrates how self-similar patches can be used to fine-tune the network parameters and improve the performance of conventional inpainting networks.
In Figure~\ref{fig:patchrecurrence}(a), we first obtain an initial inpainting result $I_{pred}$ by applying the fully pre-trained EdgeConnect~\cite{nazeri2019edgeconnect} network $f_{\theta}$ to an input image $\tilde{I}$ distorted with a squared mask.
Next, in Figure~\ref{fig:patchrecurrence}(b), we then generate a newly masked image $\tilde{I}_{pred}$ to utilize repeating structures of the given input image for inpainting by removing a self-similar patch in $I_{pred}$ (red squared region).
Then, by minimizing the MSE between $I_{pred}$ and restored image $f_{\theta}(\tilde{I}_{pred})$ from the newly masked image $\tilde{I}_{pred}$, we can fine-tune the network parameters $\theta$ to $\theta^*$.
Accordingly, we can update network parameters to the input image and can improve the performance of the inpainting network.
Note that we adapt network parameters without using the ground-truth image during the fine-tuning.
In Figure \ref{fig:patchrecurrence}(c), we show that we can also fine-tune the network by corrupting the initially restored image with random masks without explicit patch-match to find self-similar patches.
Specifically, we generate multiple training images for fine-tuning by corrupting $I_{pred}$ using random masks, then fine-tune the network by minimizing MSE between the initial inpainting result $I_{pred}$ and the restored image $f_{\theta}(A_i)$ from the newly corrupted image $A_i$.
By doing so, we demonstrate that we can adapt network parameters without explicitly searching for self-similar patches within the input as in Figure \ref{fig:patchrecurrence}(b).

Traditional methods require the use of dedicated algorithms and specialized network modules to search similar patches explicitly~\cite{barnes2009patchmatch, Darabi12:ImageMelding12, hays2007scene, huang2014image}.
In contrast, we utilize the self-similar patches by using the randomized masking (corruption) scheme during the test phase without changing the original network architecture.
Moreover, ours does not require any additional algorithm and/or network modules to find similar patches (e.g., patch-match).
As the area of similar patches can be randomly masked a few times during the randomized masking scheme, these self-similar patches can be naturally exposed to the network during the fine-tuning stage multiple times.
Thus, the inpainting network can learn to restore the missing area using the information from the similar patches, as shown by the result of Figure~\ref{fig:patchrecurrence}(c), which shows similar improvements to Figure~\ref{fig:patchrecurrence}(b).
Another advantage of this approach is that the randomized-masking-based fine-tuning mechanism does not restrict the shape, size, and number of the recurring patches.

\begin{algorithm}
	\textbf{Input:} input masked image $\tilde{I}$, original mask $M$
	
	\textbf{Require:} inpainting network $f$ and the pre-trained parameter $\theta_0$,   number of training $T$, random masks $\{M_{i}\}$,
	learning rate $\alpha$
	
	\textbf{Output:} enhanced inpainting result $f_{\theta^*}(\tilde{I}, M)$

	\nl $i$ $\leftarrow$ 0
	
	\nl $\theta \leftarrow \theta_0$
	
	\nl $I_{pred} \leftarrow f_{\theta_0}(\tilde{I}, M)$
	
	\While{i $<$ T}{
	    
	    \nl $\tilde{I}_{pred} \leftarrow I_{pred}\odot \left ( 1 - M_i \right )$
	    
	    \nl $I_{pred\left (\theta \right )} \leftarrow f_{\theta}(\tilde{I}_{pred}, M_{i})$
	    
	    \texttt{\\}

		\tcp{\small Random transformations can be applied for the loss.}
		\nl $loss_{rec}(\theta) \leftarrow \|I_{pred}\odot(1-M)-I_{pred(\theta)}\odot(1-M)\|^2$
		
		\texttt{\\}
		
		\nl $loss_{adv}(\theta) \leftarrow \text{\small VGG and/or adversarial~losses}$
		
		\nl $Loss(\theta) \leftarrow loss_{rec}(\theta) + loss_{adv}(\theta)$
		
        \texttt{\\}
		
		\tcp{\small Parameter update}
		\nl $\theta \leftarrow \theta - \alpha \nabla_{\theta}  \textit{Loss}(\theta)$

		\nl $i$ $\leftarrow $ $i$ + 1  
	}

	\nl $\theta^* \leftarrow \theta$

	\textbf{Return:}  $f_{\theta^*}(\tilde{I}, M)$
	\caption{\small \textit{Fine-tuning algorithm}}
	\label{algorithm}
\end{algorithm}

\begin{figure*}[h]
    \centering
    \includegraphics[width=0.95\textwidth]{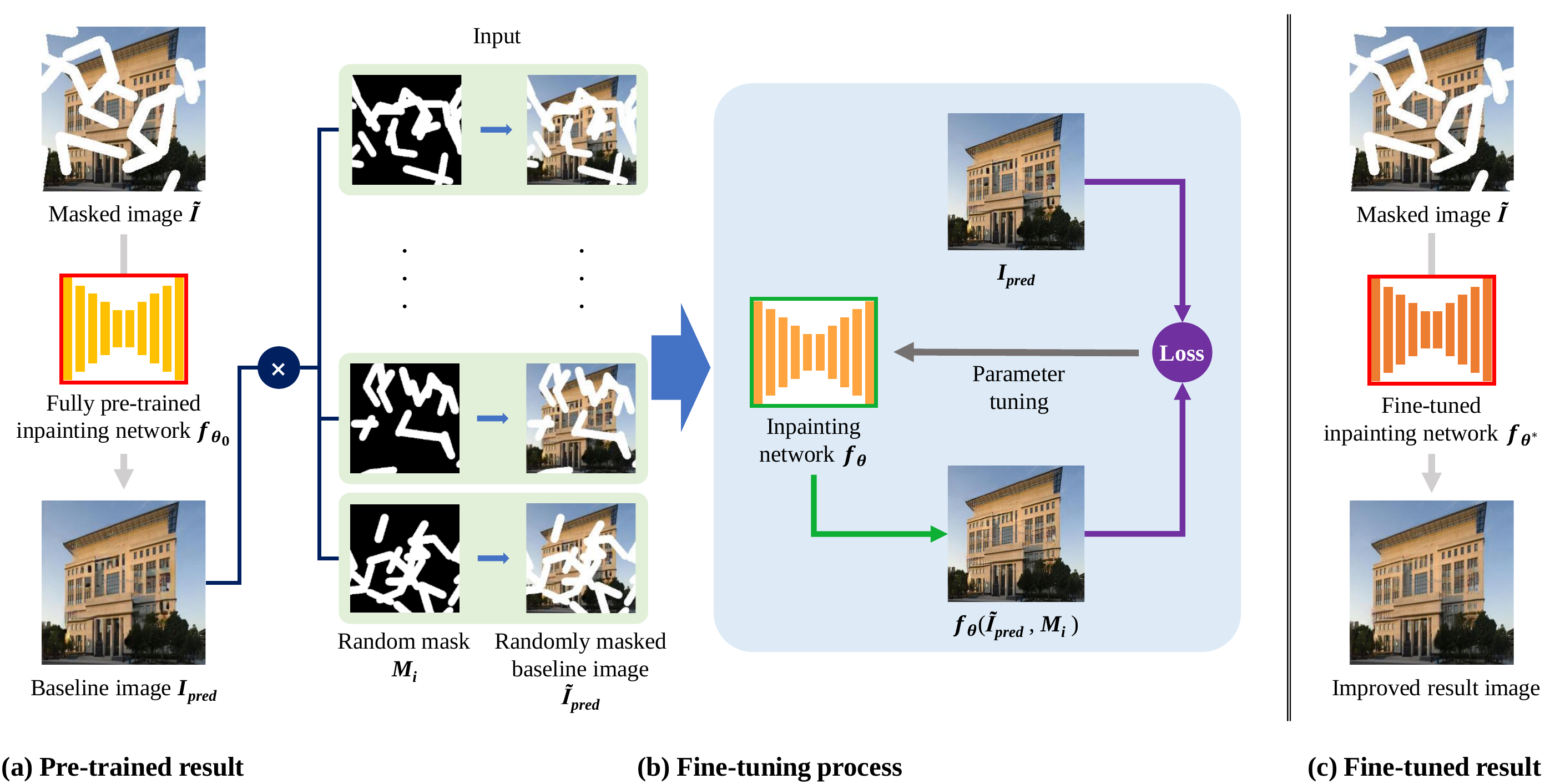}
    \caption{\small Overall flow of the proposed fine-tuning mechanism.
    (a) Initial inpainting result from the masked input image using the pre-trained network.
    (b) Fine-tuning with random masks on the result from the baseline network.
    (c) Fine-tuned network produces improved result consistent with the input image.}
    \label{fig:process}
\end{figure*}
\subsection{Overall flow}\label{overall_flow}

The overall flow of the proposed fine-tuning algorithm is described in Algorithm~\ref{algorithm} and illustrated in Figure~\ref{fig:process}.

First, we start the learning (fine-tuning) process with the initially restored image $I_{pred}$ similar to~\cite{lee2020restore} by using the fully pre-trained inpainting network as follows:
\begin{equation}
    I_{pred} = f_{\theta_0} (\tilde{I}, M ),
    \label{eq3}
\end{equation}
where $\theta_0$ denotes the fully pre-trained parameters of the baseline inpainting network $f$ and $\tilde{I}$ denotes the masked input image.
A binary map $M$ denotes a given input mask that corresponds to the input image, where missing and other areas are represented by 1 and 0, respectively.

Second, we acquire a training dataset using the initially restored image $I_{pred}$.
We generate a new and randomly corrupted masked image at the $i_{th}$ fine-tuning iteration as follows:
\begin{equation}
    \tilde{I}_{pred} = I_{pred}\odot \left ( 1 - M_i \right ),
    \label{eq5}
\end{equation}
where $M_i$ denotes the randomly generated binary mask and $\odot$ represents the element-wise multiplication.
The newly synthesized masked image $\tilde{I}_{pred}$ and initially restored image $I_{pred}$ become the input and target of our training dataset, respectively.
We render a restored image $I_{pred\left (\theta \right )}$ from the newly masked image $\tilde{I}_{pred}$ using the inpainting network $f_{\theta}$.

Third, we compute gradient values with respect to network parameters using the predefined loss functions for pre-training the baseline inpainting network and then update network parameters using a conventional optimizer (e.g., ADAM).
The loss is calculated using the difference between $I_{pred}$ and $I_{pred\left (\theta \right )}$.
Notably, if the baseline inpainting network includes GAN architecture, the discriminator of GAN architecture computes the adversarial loss by using the initially restored image $I_{pred}$ as the real sample rather than the ground-truth clean image.

Specifically, for the pixel-wise loss (i.e., reconstruction loss), we can use the L2 loss function as follows:
\begin{equation}
    loss_{rec}(\theta) = \|I_{pred}\odot(1-M)- I_{pred(\theta)}\odot(1-M)\|^2.
    \label{ExcludeMask}
\end{equation}
In practice, to compute the reconstruction loss, we exclude the part corresponding to the original mask $M$ (i.e., initially restored area) since we can improve the performance by making this slight modification. Note that, we can use any conventional reconstruction losses beyond the L2 loss (e.g., L1).
Moreover, when considering additional (perceptual) losses, such as VGG and adversarial losses, we do not need to make any modification. We employ the original function used to train the baseline inpainting networks, 
and the expression is $loss_{adv}(\theta)$.
Thus, overall loss function that updates parameters of the given inpainting network is expressed as follows:
\begin{equation}
    Loss\left ( \theta \right ) = loss_{rec}(\theta) + loss_{adv}(\theta).
    \label{OverallLoss}
\end{equation}

Then, we repeat these steps $T$ times, which is the number of iterations determined experimentally to achieve best results. Notably, random transformations can be applied during fine-tuning to prevent the network from predicting $I_{pred}$ regardless of its input after adaptation.

Finally, the network parameter is upgraded to $\theta^*$ and we obtain the final fine-tuned image ${f}_{\theta^*}(\tilde{I}, M)$ with the original input image $\tilde{I}$ and mask $M$.

We call this algorithm ``restore-from-restored" because this method leverages the already restored image to improve the inpainting performance.

\section{Experimental result}
Please refer to our supplementary material for more results. 
Moreover, the code, dataset, and pre-trained models for the experiments will be included in the supplementary material upon acceptance.
\subsection{Implementation details}
To evaluate the performance of the proposed fine-tuning algorithm, we use three different inpainting networks, namely, GatedConv~\cite{yu2019free}, EdgeConnect~\cite{nazeri2019edgeconnect}, and GMCNN~\cite{wang2018image}, as baseline networks of our algorithm.
The experiment is performed using the official code for each model.
Notably, EdgeConnect is currently a state-of-the-art inpainting network.

For our experiments, we use officially available and fully pre-trained network parameters on the Places2 dataset~\cite{zhou2017places} for each network and fine-tune the pre-trained parameters via the proposed approach in Algorithm.~\ref{algorithm}.
We evaluate the performance of the proposed algorithm on test sets in conventional benchmark datasets, such as Places2 ~\cite{zhou2017places} and Urban100~\cite{huang2015single}.
The Places2 dataset is a mixed dataset of people, landscapes, and buildings; the Urban100 dataset mainly consists of buildings with repetitive patterns.

Our random free-form masks used in the fine-tuning process are obtained by the algorithm for sampling free-form masks introduced in \cite{yu2019free}.
The input image size for the fine-tuning is 256$\times$256 and each input image includes holes covering approximately 20\% to 40\% of the image.
We utilize the same optimizer and loss function used in each of the pre-trained models, but we modify the learning rate for the fine-tuning: we use learning rate of $10^{-5}$ for EdgeConnect and GatedConv, and $10^{-4}$ for GMCNN.

Our experiments are conducted with Intel i9 and NVIDIA RTX2080 Ti GPU, and it takes 0.2, 0.3, and 0.7 seconds to perform 1 fine-tuning iteration (patch-size = 256$\times$256, batch-size = 4) with GMCNN, GatedConv, and EdgeConnect respectively.

\subsection{Ablation study}\label{sec_ablation}

\begin{figure}[t]
    \centering
    \includegraphics[width=1\linewidth]{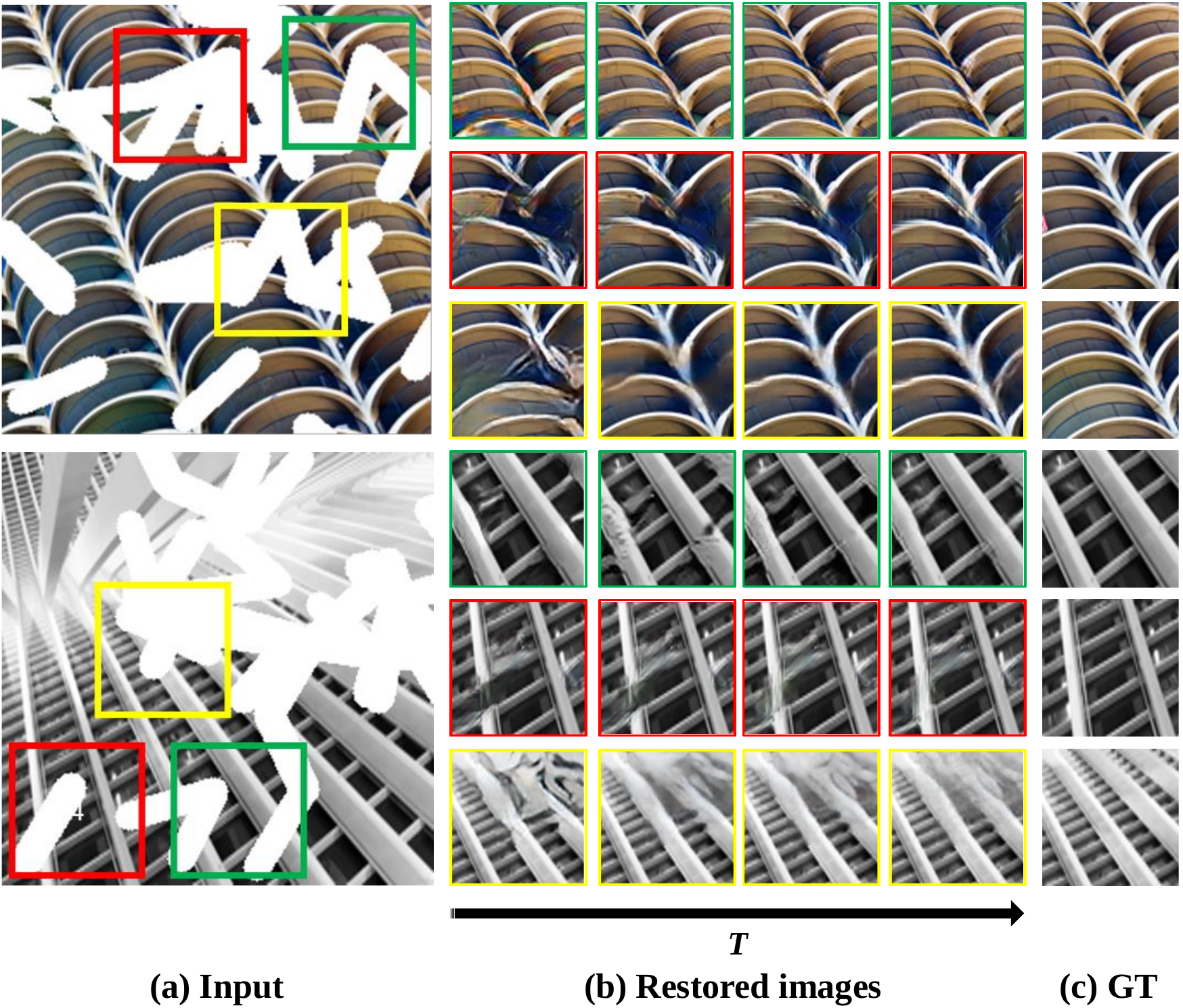}
    \caption{\small Visual results on the Urban100 dataset according to the fine-tuning progress.
    Green, red, and yellow boxes represent the results by GatedConv, EdgeConnect, and GMCNN, respectively.
    (a) Input masked images.
    (b) Initially restored image and fine-tuned images for $T$ iterations. 
    \textbf{Green box}: GatedConv results (From left to right: $T$=0, $T$=200, $T$=400, $T$=700). \textbf{Red box}: EdgeConnect results (From left to right: $T$=0, $T$=100, $T$=500, $T$=1000).
    \textbf{Yellow box}: GMCNN results (From left to right: $T$=0, $T$=50, $T$=150, $T$=250).
    (c) Ground-truth images.}
    \label{fig:iterprogress}
\end{figure}
\begin{figure}[]
    \centering
    \includegraphics[width=1\linewidth]{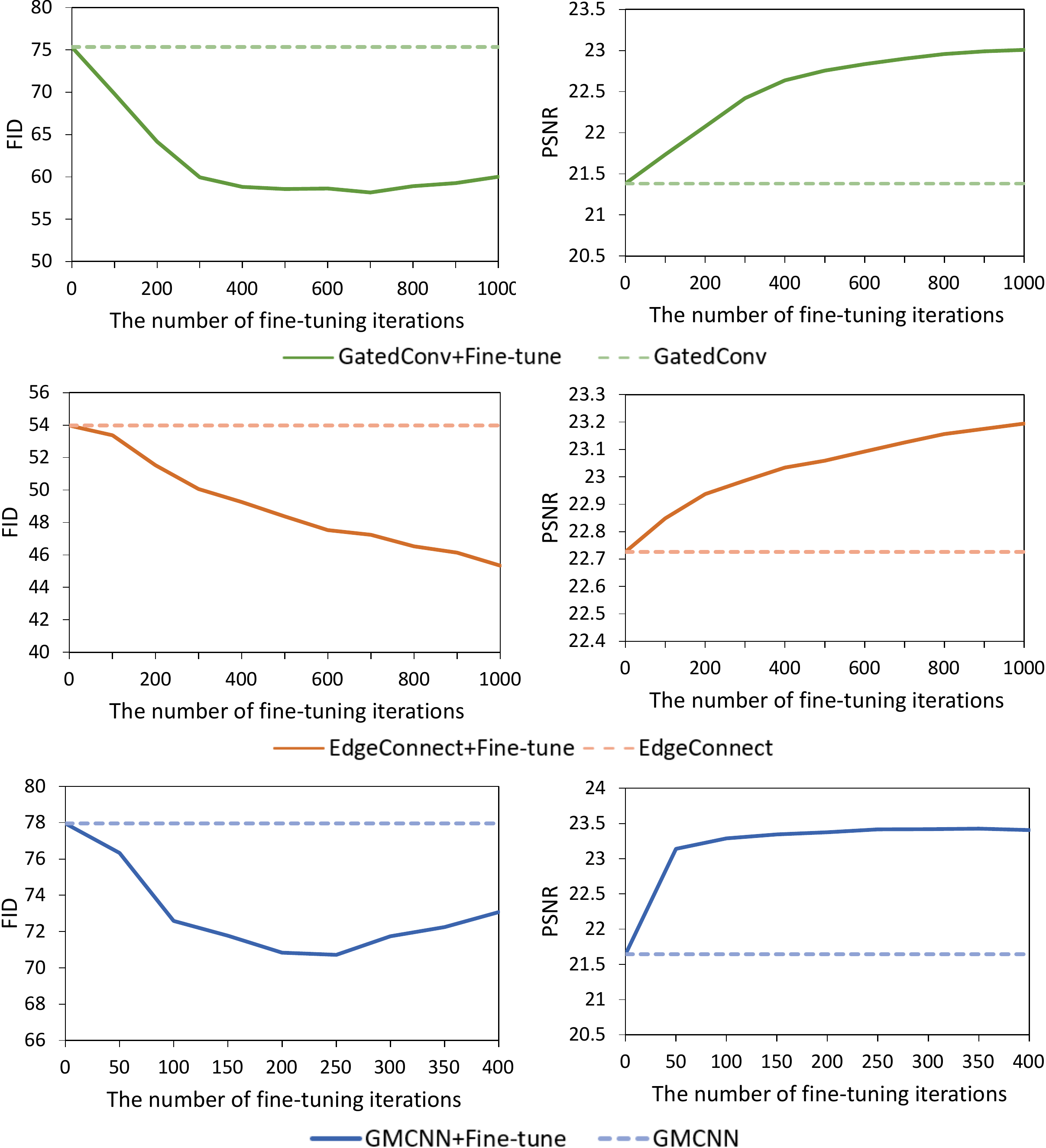}
    \caption{\small Improvement of FID and PSNR values of GatedConv, EdgeConnect, and GMCNN with the fine-tuning process on the Urban100 dataset.
    High PSNR values and low FID scores indicate better quality of images.}
    \label{fig:urbangraph}
\end{figure}

\paragraph{Number of fine-tuning iterations and image quality.}
We show changes in the image quality depending on the number of fine-tuning iterations through the experiment conducted with the Urban100 dataset.

Figure \ref{fig:iterprogress} illustrates the results of the fine-tuned network with different numbers of iterations for each pre-trained model.
The structure of the results is gradually improved as the fine-tuning progress, and the results become close to the ground-truth when it comes to the pre-fixed iteration.

Moreover, Figure \ref{fig:urbangraph} shows the trend of Fréchet inception distance (FID) scores and peak SNR (PSNR) values with a fine-tuning process for each model.
We observe that a large number of iterations for the fine-tuning can lead to over-smoothed results depending on the model, and we demonstrate that the FID score can be an important criterion for determining the proper number of iterations $T$ because of its higher sensitivity to over-smoothing than the PSNR value.

In the experiments of the remaining sections, we use the fixed number of iterations chosen in this manner for fine-tuning.
Specifically, the number of fine-tuning iterations is determined after a small random sample experiment performed on each dataset and model.
Notably, these datasets used to determine $T$ are excluded in our evaluation.
As a result, we fine-tune 500, 1000, and 200 iterations on the Places2 test dataset and 700, 1000, and 250 iterations on the Urban100 dataset for each pre-trained model (i.e., GatedConv, EdgeConnect, and GMCNN) at the test phase.

\subsection{Quantitative results}

\begin{table}[t]
\begin{adjustbox}{max width=1.0\linewidth,center}
\centering
\begin{tabular}{ccccccc}
\hline
\textbf{Model} & \textbf{\# iter.} (=$T$) & \textbf{PSNR} & \textbf{SSIM} & \textbf{L1} (\%) & \textbf{FID} & \textbf{LPIPS} \\ \hline
\multirow{5}{*}{\textbf{\begin{tabular}[c]{@{}c@{}}Gated\\ Conv\end{tabular}}} & 0 (baseline) & 23.25 & 0.8831 & 2.67 & 60.75 & 0.103 \\
 & 100 & 23.10 & 0.8767 & 2.75 & 63.47 & 0.109 \\
 & 200 & 23.36 & 0.8820 & 2.69 & 60.70 & 0.103 \\
 & 500 & 23.84 & 0.8900 & 2.60 & \textbf{59.70} & \textbf{0.097} \\
 & 1000 & \textbf{23.97} & \textbf{0.8913} & \textbf{2.60} & 63.51 & 0.101 \\ \hline
\multirow{5}{*}{\textbf{\begin{tabular}[c]{@{}c@{}}Edge\\ Connect\end{tabular}}} & 0 (baseline) & 24.00 & 0.8934 & 2.84 & 28.06 & 0.098 \\
 & 100 & 24.04 & 0.8941 & 2.83 & 27.94 & 0.098 \\
 & 200 & 24.06 & 0.8946 & 2.82 & 27.78 & 0.097 \\
 & 500 & 24.09 & 0.8952 & 2.81 & 27.53 & 0.097 \\
 & 1000 & \textbf{24.12} & \textbf{0.8957} & \textbf{2.80} & \textbf{27.23} & \textbf{0.096} \\ \hline
\multirow{5}{*}{\textbf{GMCNN}} & 0 (baseline)& 23.79 & 0.8477 & 6.02 & 69.92 & 0.078 \\
 & 100 & \textbf{24.70} & \textbf{0.8586} & \textbf{5.60} & 65.56 & 0.077 \\
 & 200 & 24.61 & 0.8550 & 5.74 & \textbf{61.60} & \textbf{0.069} \\
 & 500 & 24.41 & 0.8475 & 6.05 & 69.57 & 0.073 \\
 & 1000 & 24.18 & 0.8439 & 6.31 & 75.13 & 0.082 \\ \hline
\end{tabular}
\end{adjustbox}
\caption{\small Fine-tuning results of various inpainting models on the Places2 dataset by changing the number of iterations. PSNR, SSIM, L1, FID, and LPIPS values are measured.}
\label{tab:places2-table}
\end{table}

\begin{table}[t]
\begin{adjustbox}{max width=1.0\linewidth,center}
\centering
\begin{tabular}{cccc}
\hline
\textbf{} & \begin{tabular}[c]{@{}c@{}}\textbf{GatedConv}\\ ($T$=700)\end{tabular} & \begin{tabular}[c]{@{}c@{}}\textbf{EdgeConnect}\\ ($T$=1000)\end{tabular} & \begin{tabular}[c]{@{}c@{}}\textbf{GMCNN}\\ ($T$=250)\end{tabular} \\ \hline
\textbf{PSNR$^\star$} & 21.38 $\rightarrow$ 22.90 & 22.73 $\rightarrow$ 23.19 & 21.64 $\rightarrow$ 23.42 \\ \hline
\textbf{SSIM$^\star$} & 0.862 $\rightarrow$ 0.893 & 0.880 $\rightarrow$ 0.892 & 0.830 $\rightarrow$ 0.850 \\ \hline
\textbf{FID$^\dagger$} & 75.34 $\rightarrow$ 58.15 & 53.98 $\rightarrow$ 45.34 & 77.96 $\rightarrow$ 70.71 \\ \hline
\textbf{LPIPS$^\dagger$} & 0.096 $\rightarrow$ 0.078 & 0.085 $\rightarrow$ 0.073 & 0.083 $\rightarrow$ 0.082 \\ \hline
\end{tabular}
\end{adjustbox}
\caption{\small Changes of PSNR, SSIM, FID, and LPIPS values before and after fine-tuning. GatedConv, EdgeConnect, and GMCNN are fined-tuned on the Urban100 test dataset. The fine-tuning iteration (T) is determined empirically (refer to Sec.~\ref{sec_ablation}).
Lower $^\dagger$ and higher $^\star$ scores indicate better quality of images.}
\label{tab:urban-table}
\end{table}

We quantitatively evaluate the performance of fine-tuned networks with L1, peak signal-to-noise ratio (PSNR), and structural similarity index measure (SSIM) to compare the inpainting results objectively (L1 values are provided because it is a common metric in many inpainting studies~\cite{pathak2016context, nazeri2019edgeconnect, yu2019free}.).
These methods mainly measure the distortion of the results, assuming that the ideal results are the same as the original.

Moreover, we compare the perceptual quality of the results by calculating how plausible the produced image looks from the human perspective.
This comparison is necessary because the inpainting task does not simply recover images quantitatively but furthermore produces visually appealing images to humans based on the given input image. 
We use learned perceptual image patch similarity (LPIPS)~\cite{zhang2018unreasonable} and Fréchet inception distance (FID)~\cite{heusel2017gans} values for the perceptual comparison.
To be specific, LPIPS compares the deep features of images extracted from the image classifier based on a CNN architecture, such as VGG~\cite{simonyan2014vgg} and AlexNet~\cite{krizhevsky2014alex}.
We use AlexNet in this study.
FID measures the Wasserstein distance (Fréchet distance) between distributions of features from the real and generated images.
The mean and covariance of each distribution are calculated through the fully pre-trained Inception-V3~\cite{szegedy2016rethinking} network.

Table~\ref{tab:places2-table} and Table~\ref{tab:urban-table} list the quantitative restoration results.
First, Table~\ref{tab:places2-table} shows the comparison of the results from each fine-tuning iteration on the Places2 dataset.
One thousand images from the Places2 test dataset are used for the evaluation.
With the aid of our fine-tuning algorithm, overall metrics values improve consistently compared with the restoration results by the pre-trained baseline models (i.e., $T$=0).
Next, Table \ref{tab:urban-table} presents the improvements between the results of pre-trained models and the findings after fine-tuning on the Urban100 test dataset.
Although the initial results from the baseline models are poor since the Urban100 dataset is not used in pre-training the baseline networks, our fine-tuning results show considerable improvements.
This finding proves that our fine-tuning method enhances the results although the input test image has a slightly different distribution from the dataset used in pre-training.

\subsection{Qualitative results}
\begin{figure*}[h]
    \centering
    \begin{adjustbox}{max width=1.0\linewidth,center}
    \includegraphics{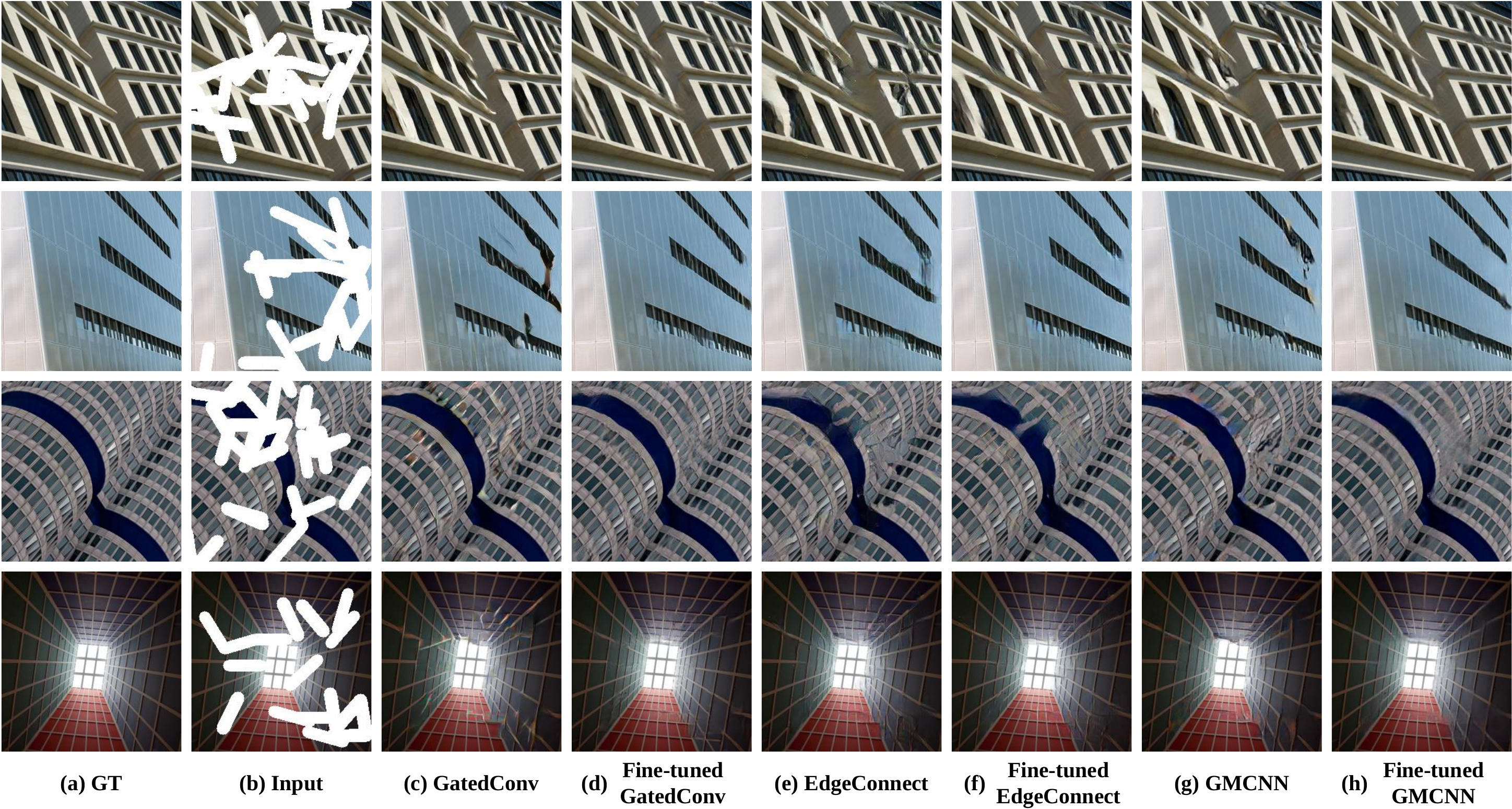}
    \end{adjustbox}
    \caption{\small Qualitative comparison of results on the Urban100 dataset before and after fine-tuning pre-trained models.
    The first two columns show the ground-truth and masked input image.
    The next two columns represent image pairs, where the left is the baseline image obtained via the pre-trained inpainting models and the right is the result images by our fine-tuning algorithm.}
    \label{fig:qualitative_urban}
\end{figure*}

\begin{figure}[h]
    \centering
    \begin{adjustbox}{max width=1.0\linewidth,center}
    \includegraphics{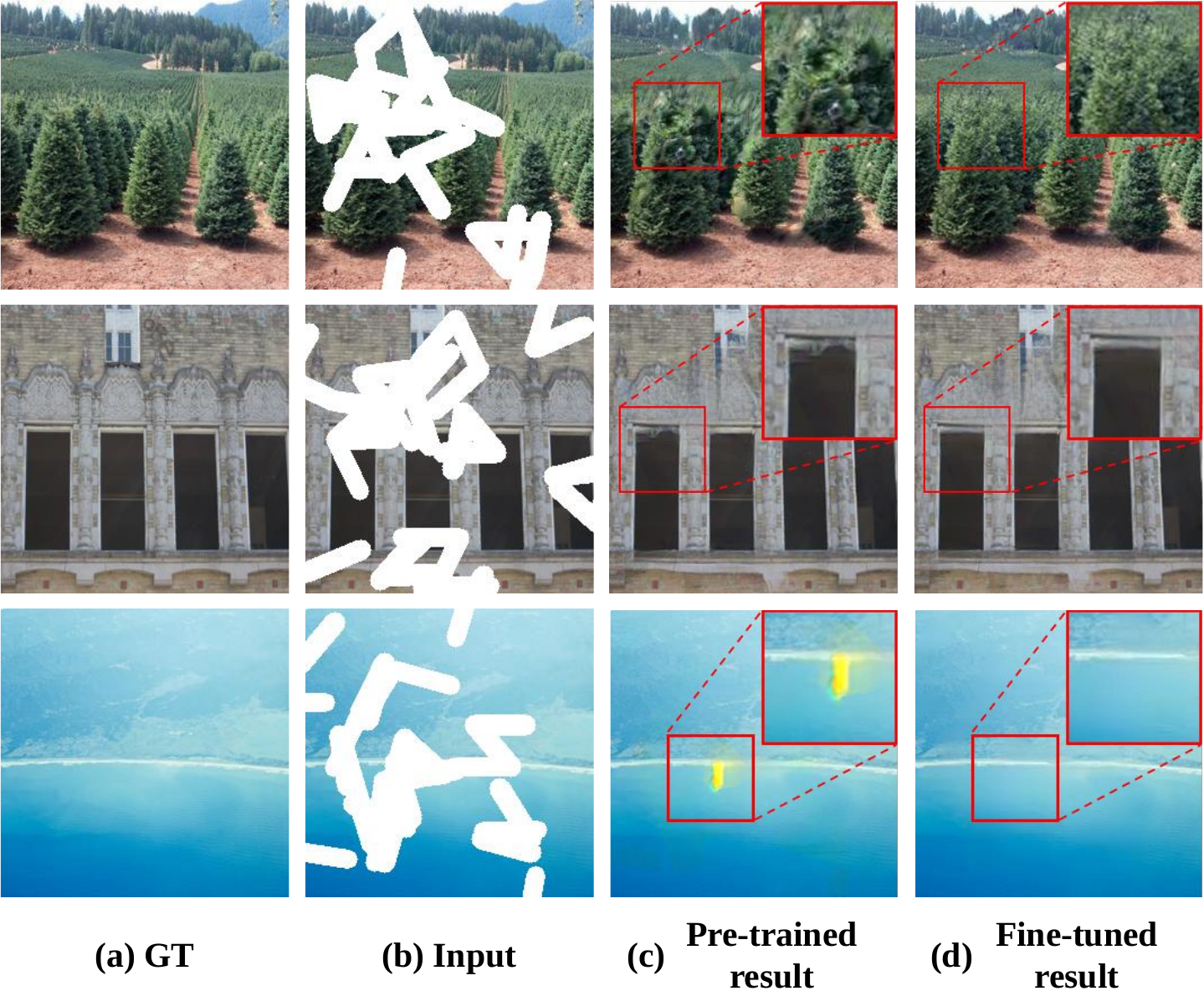}
    \end{adjustbox}
    \caption{\small Qualitative comparison of results on the Places2 dataset before and after fine-tuning the pre-trained models.
    A specific part of each result image is magnified to demonstrate improved detail.
    }
    \label{fig:qualitative_places}
\end{figure}

We compare the qualitative results between the initially restored results by the pre-trained models and our results by fine-tuning the pre-trained baselines.
Figure~\ref{fig:qualitative_places} shows visual results on the Places2 test dataset.
Note that, the generated part in the result images of pre-trained models is distorted or does not match the other part.
By comparison, our fine-tuned models generate more natural results.
Furthermore, the visual results are significantly improved if multiple repetitive patterns, such as windows and stairs, exist in the input image since the network is likely to learn the correct answer using many similar patches.
Figure~\ref{fig:qualitative_urban} shows the fine-tuning results on the Urban100 dataset.
The results on the Urban100 test dataset demonstrate improved performance for structural restoration due to the property of containing many repetitive structures within the image.

\section{Conclusion}
A new self-supervision-based inpainting algorithm that allows the adaptation of fully pre-trained network parameters during the test stage is proposed.
We utilize self-similar patches within the given input test image to fine-tune the network without using the ground-truth clean image and elevate the performance of networks by combining internal and large external information.
We can easily fine-tune the baseline networks and significantly improve the performance over the baselines by optimizing loss functions, which are used to pre-train the baseline networks.
The proposed method achieves state-of-the-art inpainting results on the conventional benchmark datasets, and extensive experimental results demonstrate the superiority of our method.

\if
\section*{Acknowledgment}
This work was supported by the research fund of Hanyang University (HY-2018).
\fi

\clearpage

{\small
\bibliographystyle{ieee_fullname}
\bibliography{inpainting}
}

\end{document}